\documentclass[letterpaper, 10 pt, conference]{ieeeconf} 

\IEEEoverridecommandlockouts                              
\def\baselinestretch{0.99}

\usepackage{graphics} 
\usepackage{epsfig} 
\usepackage{mathptmx} 
\usepackage{times} 
\usepackage{amsmath} 
\usepackage{amssymb}  
\usepackage{thm}
\usepackage{color}
\usepackage{subfig}

\newcommand{\real}{{\mathbb{R}}}
\newcommand\oprocendsymbol{\hbox{$\square$}}
\newcommand\oprocend{\relax\ifmmode\else\unskip\hfill\fi\oprocendsymbol}

\newenvironment{remark}[1][\unskip]{%
\par
\noindent
\emph{Remark #1.}
\noindent}

\usepackage{hyperref} 
\hypersetup{
    colorlinks,
    citecolor=black,
    filecolor=black,
    linkcolor=black,
    urlcolor=black,
    pdfauthor={},
    pdfsubject={},
    pdftitle={}
}

\newcommand{\AFc}[1]{{#1}}
\newcommand{\Sara}[1]{{\color{black} #1}}

\newtheorem{theorem}{Theorem}[section]

\title{\LARGE \bf From Tracking to Robust Maneuver Regulation:\\ an Easy-to-Design Approach for VTOL Aerial Robots}

\author{Sara Spedicato$^{1}$, Antonio Franchi$^{2,3}$ and Giuseppe Notarstefano$^{1}$% 
\thanks{$^{1}$Department
    of Engineering, Universit\'{a} del Salento, Via per Monteroni, 73100 Lecce,
    Italy, {\tt\scriptsize sara.spedicato@unisalento.it,
      giuseppe.notarstefano@unisalento.it}}%
  \thanks{$^2$CNRS, LAAS, 7 avenue du colonel Roche, F-31400 Toulouse, France}%
  \thanks{\hspace{-1em}$^3$Univ de Toulouse, LAAS, F-31400 Toulouse, France {\tt
      \scriptsize \href{mailto:afranchi@laas.fr}{afranchi@laas.fr}}}%
  \thanks{This work has been funded by the European Union's Horizon 2020
    research and innovation programme under grant agreement No 644271 AEROARMS
    and under European Research Council (ERC) grant agreement No 638992
    OPT4SMART.} }%

\begin{document}

\maketitle

\begin{abstract}
  In this paper we present a maneuver regulation scheme for Vertical Take-Off
  and Landing (VTOL) micro aerial vehicles (MAV). Differently from standard
  trajectory tracking, maneuver regulation has an intrinsic robustness due to
  the fact that the vehicle is not required to chase a virtual target, but just
  to stay on a (properly designed) desired path with a given velocity
  profile. In this paper we show how a robust maneuver regulation controller can
  be easily designed by converting an existing tracking scheme. The resulting
  maneuvering controller has three main appealing features, namely it: (i)
  inherits the robustness properties of the tracking controller, (ii) gains the
  appealing features of maneuver regulation, and (iii) does not need any
  additional tuning with respect to the tracking controller.  We prove the
  correctness of the proposed scheme and show its effectiveness in
experiments on a nano-quadrotor.  In particular, we show on
  a nontrivial maneuver how external disturbances acting on the quadrotor cause
  instabilities in the standard tracking, while marginally affect the maneuver
  regulation scheme.
\end{abstract}

\section{INTRODUCTION}
Typical envisioned tasks for Micro Aerial Vehicles (MAVs) include surveillance,
monitoring, inspection, search and rescue operations and the realization of
advanced robotic tasks.
As a preliminary subtask, these applications essentially require MAVs to fly
along a prescribed path with a prescribed velocity profile along it.
For all these applications, ensuring an effective and robust performance of the
flight controller represents a fundamental requirement.  The majority of these
tasks are carried out outdoors exposing the vehicle to adverse atmospheric
conditions, as, e.g., unknown wind patterns, that deteriorate the motion
performance.  Furthermore, when the vehicle operates in contact with the
environment or in formation with other vehicles, it is greatly influenced by
nonlinear (often unmodeled) aerodynamics due to surrounding objects/vehicles
\cite{JD-GD-OK-VK:13}.
More recently, the robotics community is rising a growing interest in the usage
of swarms of nano aerial vehicles with respect to fewer bigger counterparts
(\cite{MF-GC-RN-RGS-LM:15, AK-DM-CP-VK:13}).  This is mainly due to their
ability to operate in tight formations in small, constrained indoor
environments.  They are also cheaper and more robust to collisions and less
safety precautions are required in their usage.
On the other hand, maneuvering is more challenging with respect to standard
sized vehicles. In fact nano vehicles are more agile and characterized by faster
dynamics. Moreover, due to their tiny and light structure, parameter errors and
external disturbances (e.g., air flows) have a stronger impact. 

Recently, since ``real world" applications require controllers able to cope with
parameter uncertainties and external disturbances, the development of robust
control techniques has risen a considerable attention in the field of autonomous
aerial vehicles.  In particular,
the presence of force disturbances (e.g., air flows due to external sources or
proximity effects) has been largely considered.
We divide the literature in two parts considering respectively the rejection of
constant and non constant disturbances.
Constant disturbances have been considered in \cite{AR-AT:11,DC-RC-CS:14}, where
adaptive position tracking control schemes requiring force disturbance
estimation are proposed. In \cite{DC-RC-CS:14} an experimental validation of the
proposed controller is presented. The desired trajectory to be followed is a
lemniscate with a constant speed of $1$ m/s and the experiment is performed with
wind disturbances arising only from an air conditioning system. The same authors
present in \cite{DC-RC-CS:15} an experimental test in which a quadrotor is
forced to hover in the slipstream of a mechanical fan.
Other works present ad-hoc controllers developed considering more realistic
conditions: near constant \cite{LC-OMA-WHC:12,2013k-AntCatRobChiFra}, time-varying disturbances
(\cite{JE-SS-HR-RL:13, PC-NM-RN-LM:14}) and even space-varying turbulent wind
fields \cite{NS-BS-DAP:13}.  A common approach is used in these works: a
disturbance model is defined and then an estimator is adopted to determine the
disturbance model parameters.
Among the works presented above, experimental tests under windy conditions are
only presented in \cite{LC-OMA-WHC:12} on a fixed wing autonomous vehicle.
It is worth noticing that the maneuvering problem in presence of disturbances is
presented as a trajectory tracking problem in (\cite{JE-SS-HR-RL:13,
  PC-NM-RN-LM:14, NS-BS-DAP:13}) and as a path following problem in
\cite{LC-OMA-WHC:12}.

The main contribution of the paper is the design and experimental validation of
a maneuver regulation approach for Vertical Take-Off and Landing aerial vehicles
(VTOLs) based on a suitable re-design of off-the-shelf trajectory tracking
controllers.  Usually, ad hoc controllers are designed for disturbance
rejection.  As highlighted in the literature review, these controllers are
characterized by a fairly high complexity (presence of a disturbance model, a
complex vehicle model, and a parameter estimation scheme), thus requiring
time-expensive activities for design, implementation and controller
tuning. Furthermore, a higher computation effort is required to the control
hardware, which has to elaborate real-time data, as opposed to ``classical"
control schemes that do not take explicitly into account the disturbances.  On
the contrary, the maneuver regulation approach \cite{SS-GN-HHB-AF:13} allows us
to avoid all these onerous aspects and preserves simplicity while ensuring
robustness.
In this paper, we take inspiration from \cite{JH-RH:95} and \cite{JH:96}.
The tracking to maneuver regulation conversion is presented in \cite{JH-RH:95}
for feedback linearizable systems, while in \cite{JH:96} for a more general
class of nonlinear systems.  We propose the ``conversion technique" for VTOL
vehicles.  We show how, and under which conditions, a stable trajectory tracking
control law for a VTOL results into stable maneuver regulation.
As a further important contribution, we present experimental tests on a
nano-quadrotor.
To the best of our knowledge, no experimental tests have been carried out in
order to ``compare" the maneuver regulation approach with the classical
trajectory tracking.
In the first experiment we highlight the robustness of the maneuver regulation
scheme when an external disturbance holds the quadrotor.
In the second experiment, the nano-quadrotor, controlled using our maneuver
regulation scheme, performs a maneuver while dragging a small
payload.
  
The paper is organized as follows. 
In Section 2 we present the VTOL model and define the trajectory tracking and
maneuver regulation tasks.  In Section 3 we illustrate our maneuver regulation
control scheme for motion control of a VTOL vehicle, developed through a
robustification of a trajectory tracking controller.  Finally, in Section 4
experimental tests are provided in order to ``compare" the trajectory tracking
and the maneuver regulation approaches and prove the effectiveness of the
proposed maneuver regulation controller.

\section{VTOL model and Maneuver Regulation Task}

\subsection{VTOL Model}

A large class of miniature VTOLs can be described by the so called
vectored-thrust dynamical model, \cite{MDH-TH-PM-CS:13},
\begin{align}
\dot{\pmb{p}} &= \pmb{\text{v}}\label{eq:state_pos}\\
\pmb{\dot{\text{v}}}& = g \pmb{e}_3 - fm^{-1} R  \pmb{e}_3 \label{eq:state_vel}\\
\dot{R}&=R \AFc{\Omega} \label{eq:state_ang}\\
\pmb{\dot{\omega}} &= -I^{-1}\Omega I \pmb{\omega} +
I^{-1} \pmb{\gamma}, \label{eq:state_omega}
\end{align}
where $\pmb{p}=[p_1 \; p_2 \; p_3]^T$ is the position of the vehicle center of
mass expressed in the inertial frame $\mathcal{F}_i$,
$\pmb{\text{v}}=[\text{v}_1 \; \text{v}_2 \; \text{v}_3]^T$ is the linear
velocity expressed in $\mathcal{F}_i$, $\pmb{\omega}=[p \; q \; r]^T$ is the
\AFc{angular velocity of the} body frame $\mathcal{F}_b$ \AFc{with respect to $\mathcal{F}_i$, expressed in $\mathcal{F}_b$}, $\Omega \in so(3)$ is
such that, for
$\pmb{\beta} \in \real^3, \Omega \pmb{\beta} = \pmb{\omega} \times
\pmb{\beta}$, and
$R \in SO(3)$ is the rotation matrix mapping vectors in $\mathcal{F}_b$ into
vectors in $\mathcal{F}_i$.  Furthermore $m \in \real^+$ is the vehicle mass,
$I \in \real^{3 \times 3}$ is the inertia matrix, $g\in \real^+$ is the gravity constant,
and $\pmb{e}_3=[ 0 \; 0 \; 1]^T$. The vehicle is controlled by the thrust
$f \in \real$ and the torques $\gamma_1, \gamma_2, \gamma_3$ such that
$\pmb{\gamma} = [\gamma_1 \; \gamma_2 \; \gamma_3]^T$.

According to a time scale separation principle between fast and slow dynamics,
equations (\ref{eq:state_pos}-\ref{eq:state_omega}) can be divided into two
subsystems: (i) the position subsystem (\ref{eq:state_pos}-\ref{eq:state_vel})
and the attitude subsystem (\ref{eq:state_ang}-\ref{eq:state_omega}). Since the
attitude dynamics is fully actuated and can be controlled by means of dynamic
inversion, we concentrate our attention on the underactuated position
subsystem. Using a parameterization of the rotation matrix $R$ with
roll-pitch-yaw angles, respectively $\varphi, \theta, \psi$,
the subsystem (\ref{eq:state_pos}-\ref{eq:state_vel}) is
\begin{eqnarray}
\left[
\begin{array}{c}
\dot{p}_1 \\
\dot{p}_2 \\
\dot{p}_3 
\end{array}
\right]
&=&
\left[
\begin{array}{c}
\text{v}_1 \\
\text{v}_2 \\
\text{v}_3 
\end{array}
\right]
\label{eq:dot_p}
\\
\left[
\begin{array}{c}
\dot{\text{v}}_1 \\
\dot{\text{v}}_2 \\
\dot{\text{v}}_3 
\end{array}
\right]
&=&
-
\left[
\begin{array}{c}
s\varphi s\psi + c\psi s\theta c\varphi \\
-s\varphi c\psi + s\psi s\theta c\varphi \\
c\varphi c\theta
\end{array}
\right]
\frac{f}{m}
+
\left[
\begin{array}{c}
0\\
0\\
g
\end{array}
\right],
\label{eq:ddot_p}
\end{eqnarray}
where, for a generic angle $\alpha \in \real$, we define
$c\alpha:= \cos(\alpha)$ and $s\alpha:= \sin(\alpha)$. Equation
\eqref{eq:ddot_p} depends on the yaw angle $\psi$ which can be controlled
independently without affecting the position maneuvering objective.  
By defining $\pmb{\Phi} = [\varphi \; \theta \; \psi]^T$, equation
\eqref{eq:state_ang} can be written as
\begin{equation}
\dot{\pmb{\Phi}} = J \; \pmb{\omega},
\label{eq:jac}
\end{equation}
where $J \in \real^{3 \times 3}$ is the Jacobian matrix, which is always
invertible out of representation singularities. By choosing  
\begin{equation}
\pmb{\omega} = J^{-1}
\pmb{\mu}_{\Phi},
\label{eq:omega}
\end{equation}
where $\pmb{\mu}_\Phi = [\mu_\varphi \; \mu_\theta \; \mu_\psi]^T$ and
$\mu_\varphi, \mu_\theta, \mu_\psi$ are additional inputs and substituting
\eqref{eq:omega} in \eqref{eq:jac}, we get for the yaw angle
\begin{equation}
\dot{\psi} = \mu_\psi.
\label{eq:dot_psi}
\end{equation}
The system (\ref{eq:dot_p}-\ref{eq:ddot_p}), together with \eqref{eq:dot_psi}, can be written in state-space form as 
\begin{equation}
\dot{\pmb{x}}(t) = f(\pmb{x}(t),\pmb{u}(t))
\label{eq:state_space}
\end{equation}
with state $\pmb{x} \in \real^7$ given by $\pmb{x} = [\pmb{p}^T \; \pmb{\text{v}}^T \psi]^T$ and input $\pmb{u} \in \real^4$ given by
$\pmb{u} = [f \; \varphi \; \theta \; \mu_\psi]^T$. 
We want to point out that the roll angle $\varphi$, the pitch angle $\theta$ and the yaw-rate $\mu_\psi$
play the role of virtual control inputs that we assume being tracked by the actual
inputs at a faster rate. This is a quite common hierarchical control scheme in
commercial \AFc{VTOLs such as, e.g.,} quadrotors.

 \subsection{Trajectory Tracking and Maneuver Regulation Tasks}
 Let $(\pmb{x}_d(\cdot),\pmb{u}_d(\cdot))$ be a desired state-control trajectory, satisfying the state-space equations, i.e.,
 $$
 \pmb{\dot{x}}_d(t)=f(\pmb{x}_d(t),\pmb{u}_d(t)),
 \quad \forall t\geq0.
 $$
 It is worth noticing that state-control trajectories for the standard vehicle
 model used in this paper can be generated by exploiting its differential
 flatness, see, e.g., \cite{mistler2001exact}.
 For more general models or in case state and input constraints need to be
 explicitly taken into account in the desired trajectory generation, nonlinear
 optimal control based trajectory-generation techniques, as the ones developed
 in \cite{GN-JH-RF:07,GN-JH:10}, may be used.  

The trajectory tracking and maneuver regulation problems are defined as follows.

\subsubsection{Trajectory tracking problem}
 
 Given the desired state-control trajectory
 $(\pmb{x}_d(\cdot),\pmb{u}_d(\cdot))$, the trajectory tracking problem for the
 system \eqref{eq:state_space} consists of finding a (stabilizing) feedback
 control law  
$
 \pmb{u} = \beta(\pmb{x},t), \; \forall t \geq 0,
$
 where $\beta : \real^n \times \real_0^+ \rightarrow \real^m$, 
 such that 
\[
 \pmb{x}(t) \rightarrow \pmb{x}_d(t) \quad \text{as} \quad t \rightarrow \infty.
\]
 
\subsubsection{Maneuver regulation problem}
\Sara{Given the desired state-control trajectory
$(\pmb{x}_d(\cdot),\pmb{u}_d(\cdot))$, we define a maneuver
$[\pmb{x}_d,\pmb{u}_d]$ as the set of all the trajectories
$(\hat{\pmb{x}}_d(\cdot), \hat{\pmb{u}}_d(\cdot))$ such that 
$\hat{\pmb{x}}_d(t)=\pmb{x}_d(\sigma(t))$ and
$\hat{\pmb{u}}_d(t)=\pmb{u}_d(\sigma(t))$, $\forall t \geq 0$, for some function $\sigma : \real^+_0 \rightarrow \real^+_0$.
Now, let a maneuver $[\pmb{x}_d,\pmb{u}_d]$ be given. The maneuver regulation
problem consists of finding a feedback control law
$\pmb{u} = \beta(\pmb{x};[\pmb{x}_d,\pmb{u}_d])$ providing exponentially stable
maneuver regulation, i.e., such that there exist $ k, \lambda > 0$ and a function $\sigma : \real^+_0 \rightarrow \real^+_0$ such
that
\[
\lim_{t\rightarrow \infty} || \pmb{x}(t) - \pmb{x}_d(\sigma(t)) || \leq k e^
{-\lambda t}.
\]}
\begin{remark}[1]
In the maneuver regulation problem a system is not assigned a desired time
law, but rather is asked to reduce the ``distance" between the current state
and the entire desired state curve.
This level of flexibility gives the
maneuver regulation an intrinsic robustness to external disturbances as
opposed to the trajectory tracking approach in which the actual state is
required to ``catch up'' a desired reference.
As a consequence, some major drawbacks (due to the requirement of catching up
the reference), such as huge acceleration peaks and poor geometric tracking
of the planned path, do not arise in maneuver regulation schemes. \oprocend
\end{remark}

\section{Maneuver Regulation VTOL Control Scheme}
In this section we present our VTOL maneuver regulation control scheme, which is
based on a suitable conversion from a trajectory tracking control law.
We first present a tracking controller that we implemented in our experimental
testbed, and then show how to convert it into a maneuver regulation controller.

Let the desired position $\pmb{p}_d(\cdot)$, 
velocity $\dot{\pmb{p}}_d(\cdot)$, acceleration $\ddot{\pmb{p}}_d(\cdot)$, yaw
angle $\psi_d(\cdot)$ and yaw rate $\dot{\psi}_d(\cdot)$, be given.
By considering the vertical dynamics, i.e., the third row of \eqref{eq:ddot_p},
\begin{equation}
m \ddot{p}_3 = m g - c\varphi c\theta f,
\label{eq:vert_dyn}
\end{equation}
we choose the \emph{thrust control} as
\begin{equation}
f = - \frac{m (\mu_3-g)}{c\varphi c\theta},
\label{eq:f_control}
\end{equation}
where $\mu_3$ is an additional input to be defined later.  
The thrust control \eqref{eq:f_control} is well defined as long as the system is
away from the singularity $c\varphi c \theta = 0$.  
Let us now consider the horizontal dynamics, i.e., the first two rows of
\eqref{eq:ddot_p}, written in the form
\begin{align}
m \left[
\begin{array}{c}
	\ddot{p}_1 \\
	\ddot{p}_2 \\
\end{array}
\right]
&=
- f \;  Q(\psi) 
\left[
\begin{array}{cc}
s\varphi \\
s\theta \; c\varphi\\
\end{array} 
\right]
\label{eq:hor_dyn1}
\end{align}  
where
\[
Q(\psi)
:=
\left[
\begin{array}{cc}
s\psi & c\psi \\
-c\psi & s\psi \\
\end{array}
\right]
\]
is invertible with inverse $Q(\psi)^{-1} = Q(\psi)^T$.
By replacing \eqref{eq:f_control} in \eqref{eq:hor_dyn1}, equations
\eqref{eq:hor_dyn1} become
\begin{align}
\left[
\begin{array}{c}
\ddot{p}_1 \\
\ddot{p}_2 \\
\end{array}
\right]
&=
(\mu_3 - g) \;  Q(\psi) 
\left[
\begin{array}{cc}
\tan\varphi/c\theta \\
\tan\theta\\
\end{array} 
\right].
\label{eq:hor_dyn2}
\end{align}
We choose the \emph{roll and pitch commands}, respectively $\varphi$ and $\theta$,
as
\begin{align}
\varphi &= \text{atan}(c\theta \; \tilde{u}_1), \label{eq:phi_c}\\
\theta &= \text{atan}(\tilde{u}_2), \label{eq:theta_c}
\end{align}
where
\begin{align*}
\left[
\begin{array}{cc}
\tilde{u}_1 \\
\tilde{u}_2 \\
\end{array}
\right]
=
\frac{1}{(\mu_3 - g)} 
Q(\psi)^{-1}
\left[
\begin{array}{c}
\mu_1 \\
\mu_2 \\
\end{array}
\right],
\label{eq:phi_theta_c}
\end{align*} 
and $\mu_1, \mu_2$ are additional inputs that will be defined later.
Substituting \eqref{eq:phi_c} and \eqref{eq:theta_c} in equation \eqref{eq:hor_dyn2}, substituting \eqref{eq:f_control} in \eqref{eq:vert_dyn} and defining $\pmb{\mu} = [\pmb{\mu}_p^T \; \mu_\psi]^T$, where $\pmb{\mu}_p = [\mu_1 \; \mu_2 \; \mu_3]^T$, we get the linear system
\begin{align*}
\ddot{\pmb{p}} &= \pmb{\mu}_p,\\
\dot{\psi} &= \mu_\psi,
\end{align*}
which can be expressed in state space form as
\begin{equation}
\dot{\pmb{z}}(t) = A \pmb{z}(t) + B \pmb{\mu}(t),
\label{eq:linear}
\end{equation}
with state $\pmb{z} = [\pmb{p}^T \; \pmb{\text{v}}^T \; \psi]^T$, input
$\pmb{\mu} \in \real^4$ and system matrices
\begin{align}
A =
\left[
\begin{array}{ccc}
0_{3 \times 3} & I_{3 \times 3} & 0\\
0_{3 \times 3} & 0_{3 \times 3} & 0\\
0 & 0 & 0
\end{array}
\right],
\quad
B =
\left[
\begin{array}{cc}
0_{3 \times 3} & 0\\
I_{3 \times 3} & 0\\
0_{3 \times 1} & 1
\end{array}
\right].
\label{eq:AB}
\end{align}
Here we have denoted with $0_{i\AFc{\times} j}$ the $i \times j$ zero matrix and with $I_{i\AFc{\times }j}$ the
$i \times j$ identity matrix.
Let us define the tracking errors, respectively $\tilde{\pmb{p}}(t) :=\pmb{p}(t) - \pmb{p}_d(t)$,
$\dot{\tilde{\pmb{p}}}(t) :=\dot{\pmb{p}}(t) - \dot{\pmb{p}}_d(t)$ and $\tilde{\psi}(t) :=\psi(t) - \psi_d(t), \; \forall t \geq 0$.
The control input 
\begin{align}
\pmb{\mu}_p(t) &= \ddot{\pmb{p}}_{d}(t) - k_p \tilde{\pmb{p}}(t) - k_d  \dot{\tilde{\pmb{p}}}(t), \label{eq:mu_p}\\
\mu_\psi(t) &= \dot{\psi}_d(t)-k_\psi \tilde{\psi}(t), 
\label{eq:mu_psi}
\end{align}
with $k_p, k_d, k_\psi$ positive constants, results into an exponentially stable tracking.
This feedback linearizing controller resembles other 
tracking
schemes proposed in the VTOL literature as, e.g., in~\cite{LD-AF-ISH-HC-HHB:13}.
Being a tracking controller, it shows the previously highlighted drawbacks,
which we propose to overcome by converting it into a maneuver regulation scheme.

We are now ready to present our maneuver regulation control law.
We take advantage of the previously designed tracking controller in order to
exponentially stabilize the origin of the maneuver regulation error dynamics
instead of the tracking error dynamics.
We define the maneuver regulation error as 
$\pmb{z}(t) - \pmb{z}_d(\bar{t})$, where $\bar{t}=\pi(\pmb{z})$
 and $\pi:\real^n \rightarrow \real^+_0$
is a projection function that selects an appropriate 
time $\bar{t}$ 
to be used for maneuver regulation, according to the actual vehicle state $\pmb{z}$. The projection function $\pi(\cdot)$ is defined as
\begin{equation}
\pi(\pmb{z}) := \text{arg} \min_\tau || \pmb{z} - \pmb{z}_d(\tau)||_P^2,
\label{eq:pi}
\end{equation}
with $P > 0$. 
\Sara{Furthermore, note that $\sigma = \pi \circ \pmb{z}$}.
Thus, in order to design our maneuver regulation control law, instead of using \eqref{eq:mu_p} and \eqref{eq:mu_psi}, we choose
\begin{align}
\pmb{\mu}_p &= \ddot{\pmb{p}}_{d}(\pi(\pmb{z})) - k_p \left(\pmb{p} -
              \pmb{p}_d(\pi(\pmb{z})) \right) - k_d  \left( \dot{\pmb{p}} -
              \dot{\pmb{p}}_d(\pi(\pmb{z})) \right) \label{eq:mu_p_manreg}\\
\mu_\psi &= \dot{\psi}_d(\pi(\pmb{z}))-k_\psi \left( \psi - \psi_d(\pi(\pmb{z}))
           \right). 
\label{eq:mu_psi_manreg}
\end{align}
The convergence of the proposed maneuver regulation scheme is based on the following result, given in \cite{JH-RH:95}.
\begin{theorem}
    Let a linear system $\dot{\pmb{z}}= A \pmb{z} + B \pmb{\mu}$ and a desired
    trajectory $(\pmb{z}_d(\cdot),\pmb{\mu}_d(\cdot))$ be given.  Let us consider
    a control law ${\pmb{\mu}}=\pmb{\mu}_d(t) + K (\pmb{z}-\pmb{z}_d(t))$ such
    that the closed loop system $\dot{\pmb{e}}= A_c \pmb{e}$, with $A_c = A+BK$
    and $\pmb{e}=\pmb{z}-\pmb{z}_d$, provides uniform asymptotic tracking, i.e.,
    such that $\pmb{z}(t) \rightarrow \pmb{z}_d(t)$ as $t \rightarrow \infty$.
    Assume that there is a $c>0$ such that the projection mapping $\pi(\cdot)$,
    defined in \eqref{eq:pi}, is well defined on
    \[
    \Omega_c := \{ \pmb{z} \in \real^n : || \pmb{z} - \pmb{z}_d(\tau)||^2_P < c, \tau \in \real\}, 
    \]
    where $P > 0$ is such that $Q:= -(A_c^T P + P A_c) > 0$.
    Then the control law 
    \[
    \pmb{\mu} = \pmb{\mu}_d(\pi(\pmb{z})) + K (\pmb{z} - \pmb{z}_d(\pi(\pmb{z})))
    \]
    provides exponentially stable maneuver regulation.
\end{theorem}
\Sara{Note that $\Omega_c$ is a set on which the minimizing $\tau$ in \eqref{eq:pi} is unique \cite{JH-RH:95}.}

Next we discuss two main appealing features of the proposed maneuver regulation
control law: (i) robustness properties of the tracking controller are inherited
by the maneuver regulation scheme and (ii) no additional parameter tuning
is needed.

First, let us explain in what sense the maneuver regulation controller inherits
the robustness properties of the tracking controller. 
In order to reject
constant or slow varying disturbances, an integral control action is often
incorporated into trajectory tracking controllers for VTOL UAVs. As regards,
e.g., our 
testbed, the battery discharge, a wrong mass
estimation and the presence of a bias on angular measures cause significant
disturbances, which can be rejected by an integral control action on the
position dynamics subsystem.  For this reason, instead of \eqref{eq:mu_p}, we
choose
$$
\pmb{\mu}_p(t) = \ddot{\pmb{p}}_{d}(t) - k_p \tilde{\pmb{p}}(t) - k_d  \dot{\tilde{\pmb{p}}}(t) - k_i \pmb{\eta}_p(t),
$$ 
where $\pmb{\eta}_p \in \real^3$ denotes the state of the integrator
$\dot{\pmb{\eta}}_p(t) = \tilde{\pmb{p}}(t)$ and $k_i$ is a positive constant.
Once the tracking controller provides the integral control action, the maneuver
regulation controller can be chosen as
$$
\pmb{\mu}_p = \ddot{\pmb{p}}_{d}(\pi(\pmb{z})) - k_p (\pmb{p} - \pmb{p}_d(\pi(\pmb{z}))) - k_d  (\dot{\pmb{p}} - \dot{\pmb{p}}_d(\pi(\pmb{z})))- k_i  \pmb{\eta}_p,
$$
with the integrator $\dot{\pmb{\eta}}_p = \pmb{p} - \pmb{p}_d(\pi(\pmb{z}))$.
Notice that to prove the scheme with the integral control action, one just needs
to consider in \eqref{eq:AB} suitable augmented matrices $A_\eta$ and $B_\eta$
obtained by adding the integral state dynamics.

Second, as it clearly appears by comparing expressions \eqref{eq:mu_p},
\eqref{eq:mu_psi} with \eqref{eq:mu_p_manreg}, \eqref{eq:mu_psi_manreg}, for the
maneuver regulation scheme the same controller parameters computed in the
tracking scheme can be used.
When working with different VTOL robots, the controller gains $k_p, k_d, k_\psi$
in \eqref{eq:mu_p}, \eqref{eq:mu_psi}, have to be carefully tuned in order to
have a satisfactory behavior for the closed loop system. This tuning is a
time-consuming activity that is often carried out by the VTOL sellers. Thus,
having the possibility to use already tuned parameters is another appealing
feature of our maneuver regulation controller.
 
Finally, it is worth noticing that the possibility to convert a tracking
controller into a maneuver regulation scheme is not restricted to the particular
tracking controller presented in this section.  The ``conversion
procedure" can be applied to any feedback linearizing trajectory tracking
control law. In fact, Theorem III.1 just requires a stabilizing trajectory
tracking controller designed for a linear system
$\dot{\pmb{z}}= A \pmb{z} + B \pmb{\mu}$.

\section{Experimental Verification}
\label{sec:experiments}
In this subsection we present illustrative experiments in order to (i) prove the
effectiveness of the proposed maneuver regulation scheme and (ii) highlight its
advantages and robustness with respect to the trajectory tracking approach. We
invite the reader to watch the attached video related to these experiments.

\subsection{Experimental Platform}

\begin{figure}[ht]
	\centering
	\includegraphics[scale=0.7]{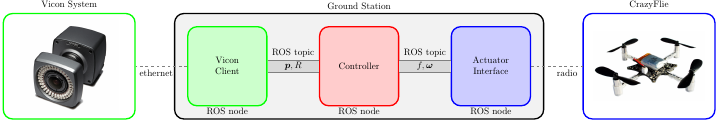}
	\caption{Hardware and ROS Nodes}
	\label{fig:architecture}
\end{figure}

To run our experiments we use a small and lightweight vehicle, belonging to the
category of nano quad-rotors, named CrazyFlie
(https://www.bitcraze.io/crazyflie/).  
Angular rates are measured on-board,
while position and attitude are measured off-board by a Vicon motion capture
system with $10$ cameras.
As regards the control architecture, a faster inner loop angular rate control
runs on-board at $500$ Hz, while the slower outer loop position/attitude control
runs at $100$ Hz on a dedicated ground station.
The ground station is equipped with our software architecture for maneuvering
control, depicted in Figure \ref{fig:architecture}, and based on a ROS
middleware.
The \emph{Vicon client node} communicates vehicle position and orientation to
the \emph{controller node}. The latter sends thrust and angular rate commands to
the \emph{actuator interface node}, which communicates with the “physical”
quadrotor through a wireless radio antenna.

\subsection{Experimental Results}
The first experimental test is as follows. We choose as desired
trajectory a circle on the horizontal plane with radius r = 0.25 m, reference
speed norm $\text{v}_d = ||\pmb{\text{v}}_d|| = 0.1 \; \text{m/s}$ and yaw angle
$\psi_d = 0$ along the curve. In order to perform the desired motion, the
quadrotor is first controlled in order to hover into a neighborhood of the
position $\pmb{p}_d(0) = [0.25 \;\; 0.0 \; -1.0 ]^T \; \text{m}$, using a
standard hovering controller. Then, we switch from the hovering task to the
circular motion task, choosing either the trajectory tracking controller or the
maneuver regulation one.
In order to test the behavior of the system in presence of exogenous
disturbances decelerating the vehicle motion and eventually immobilizing it, we
operate as follows.  During the hovering phase we hold the vehicle and switch
from the hovering task to the circular motion task. The vehicle is thus
constrained inside a neighborhood of $\pmb{p}_d(0)$ with practically zero
velocity and completely released after few seconds. This scenario is tested by
using both the trajectory tracking and the maneuver regulation schemes. The
results are depicted respectively in the left and right columns of Figure~2.
\begin{figure}[htbp]%[ht!]
	\begin{center}
		\subfloat[\emph{Path} $(p_1,p_2)$: desired position during holding time (dashed red), desired position after release (solid red), actual position (blue).]
		{\hspace{-0.3cm}\includegraphics[width=4.3cm]{{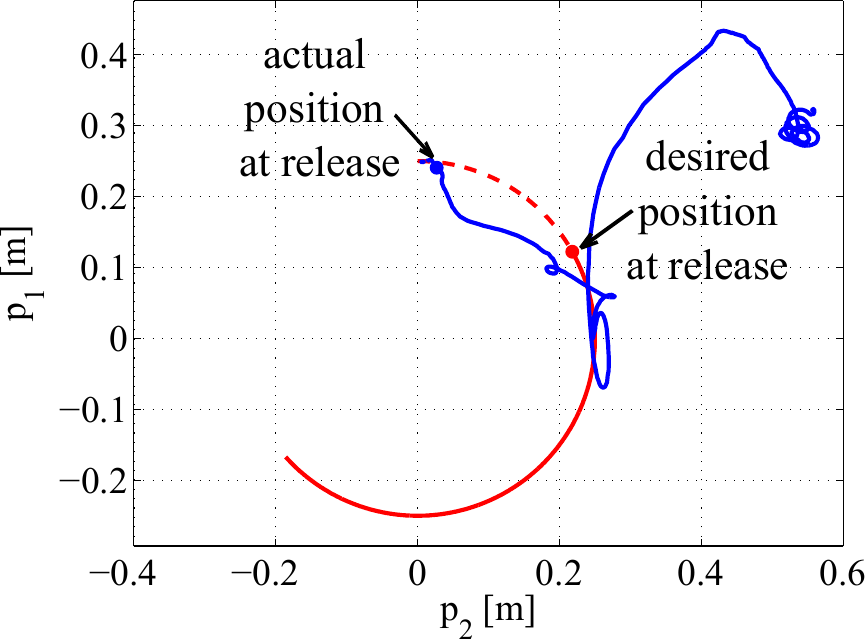}}\label{fig:xy_traj_track} \hspace{-0.1cm} 
		 \includegraphics[width=4.3cm]{{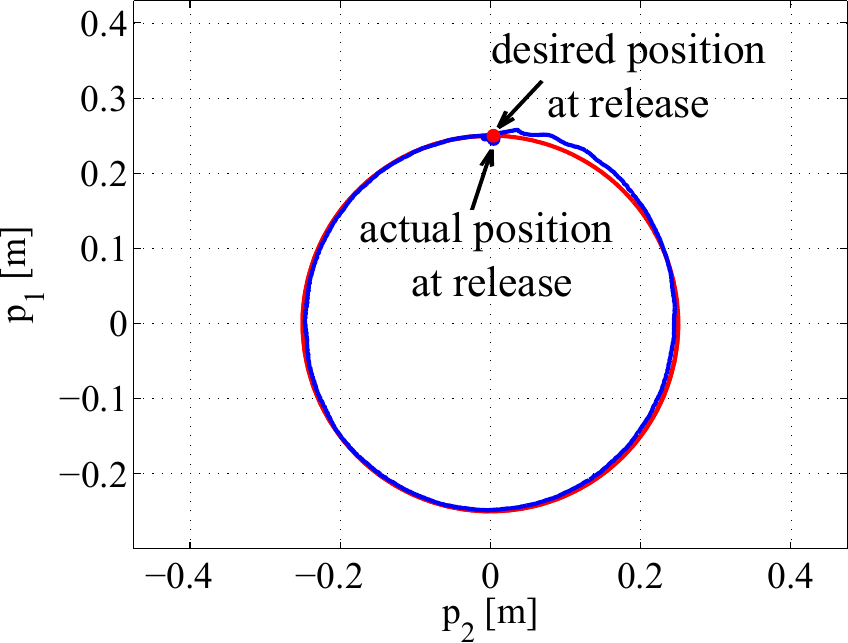}}\label{fig:xy_traj_manreg}}\\
		\vspace{-0.1cm}
		\subfloat[\emph{Positions}: $p_1(t)$ (magenta), $p_2(t)$ (black), $p_3(t)$ (green); dashed lines are desired positions, while solid lines are actual positions.] 
		{\hspace{-0.1cm}\includegraphics[width=3.9cm]{{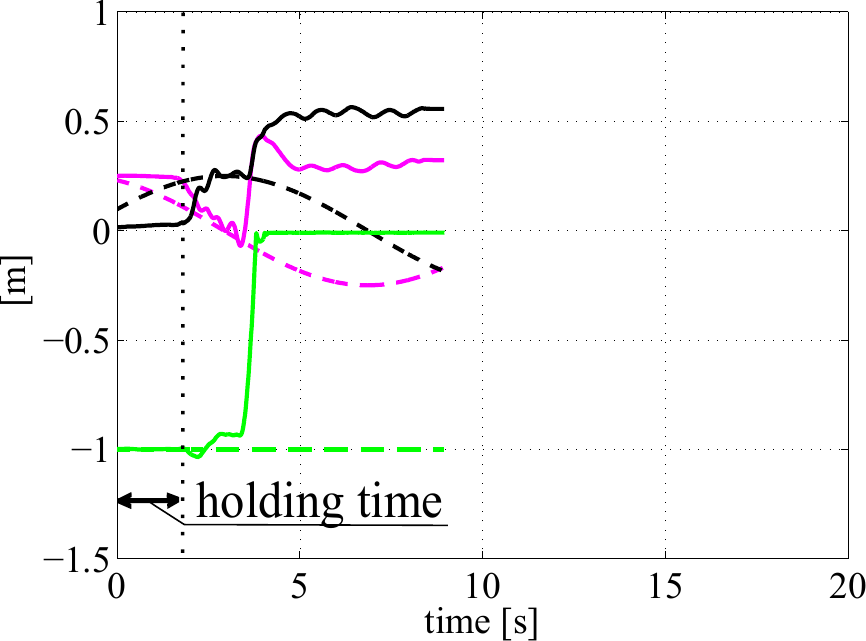}}\label{fig:p1p2p3_track}
			\hspace{0.36cm}\includegraphics[width=3.9cm]{{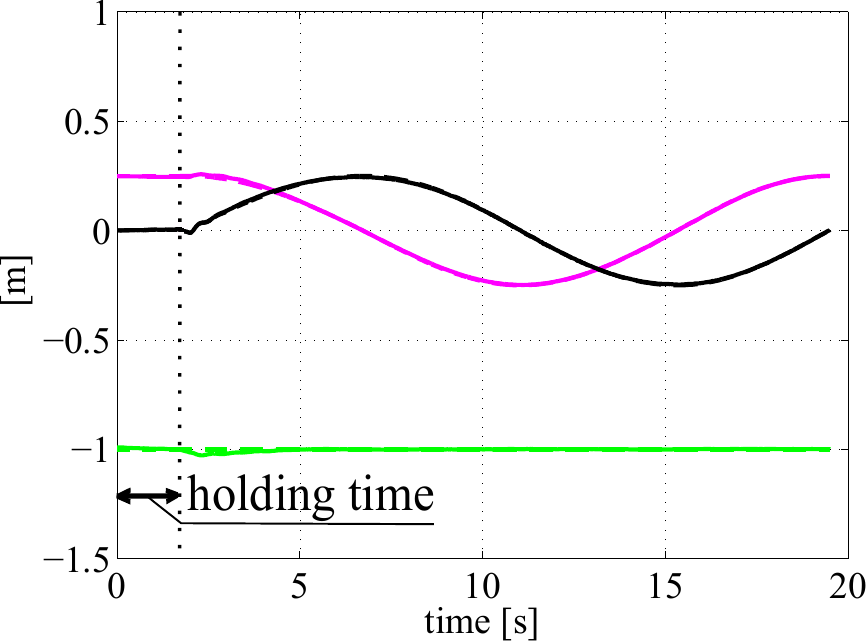}}\label{fig:p1p2p3_manreg}}\\
		\vspace{-0.1cm}
		\hspace{2.2cm}
		\subfloat[\emph{Velocity norm} $\text{v}(t)$: desired velocity (red), actual velocity (blue). 
				Note a scale factor of 10 regarding the ordinate axis.
		]
		{ \includegraphics[width=3.8cm]{{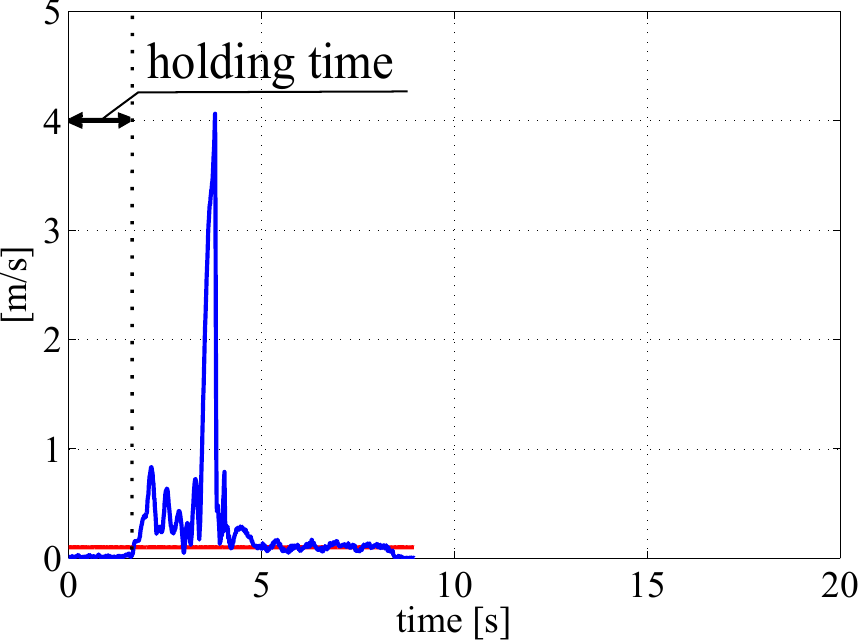}}\label{fig:vel_track}
			\hspace{0.4cm} \includegraphics[width=3.8cm]{{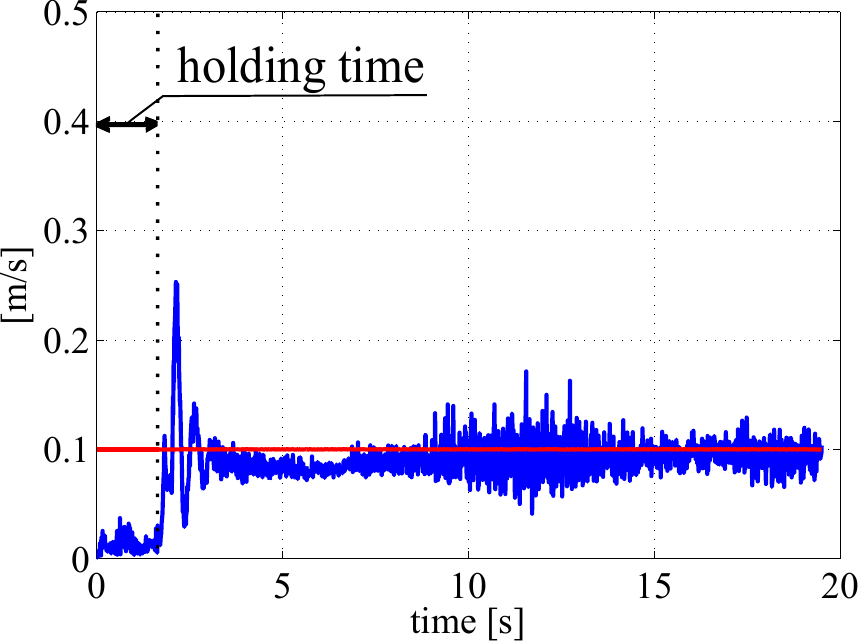}}\label{fig:vel_manreg}}\\
		\vspace{-0.1cm}
		\subfloat[\emph{Roll angle} $\varphi(t)$: roll angle command (red), actual angle (blue). 
				Note a scale factor of 10 regarding the ordinate axis.
		]
		{\hspace{-0.3cm} \includegraphics[width=4.0cm]{{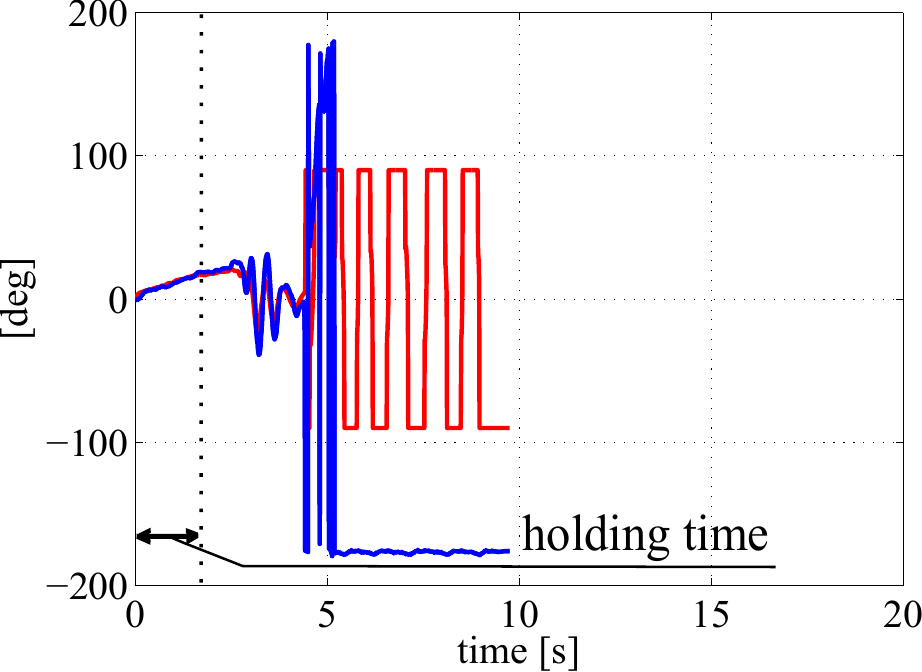}}\label{fig:phi_track}
			\hspace{0.15cm}
			\includegraphics[width=4.0cm]{{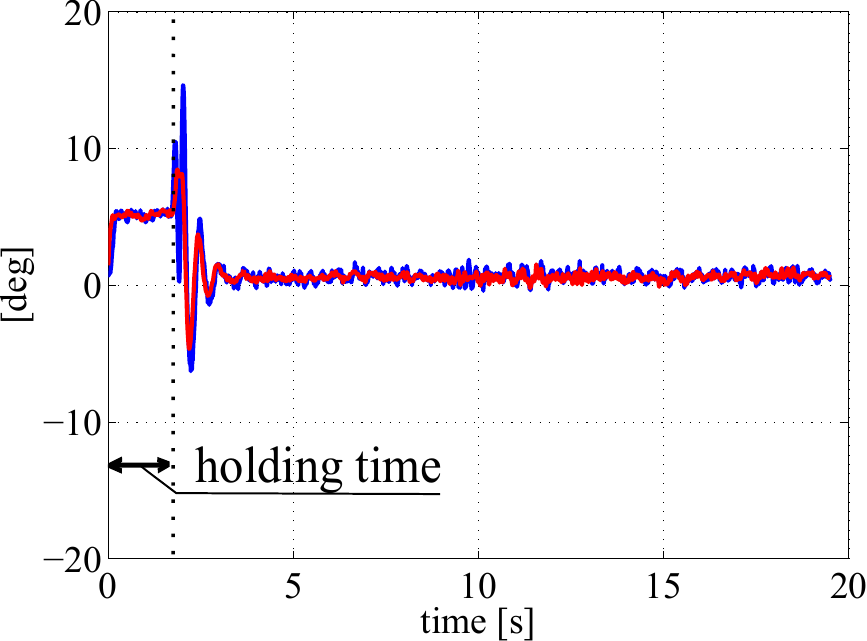}}\label{fig:phi_manreg}}\\
		\vspace{-0.1cm}
		\subfloat[\emph{Pitch angle} $\theta(t)$: pitch angle command (red), actual angle (blue). 
		Note a scale factor of 10 regarding the ordinate axis.
		]
		{\hspace{-0.3cm} \includegraphics[width=4.0cm]{{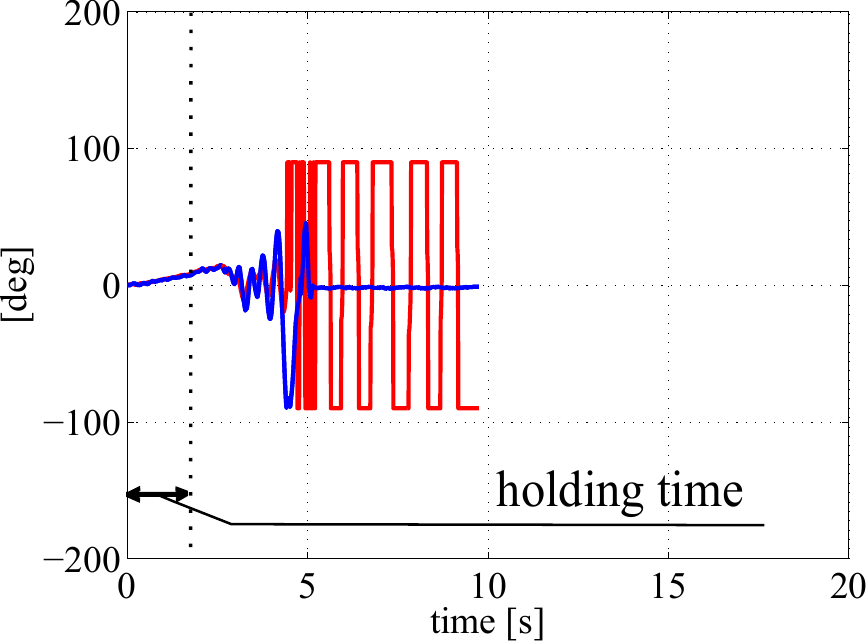}}\label{fig:theta_track}
			\hspace{0.15cm} 
			\includegraphics[width=4.0cm]{{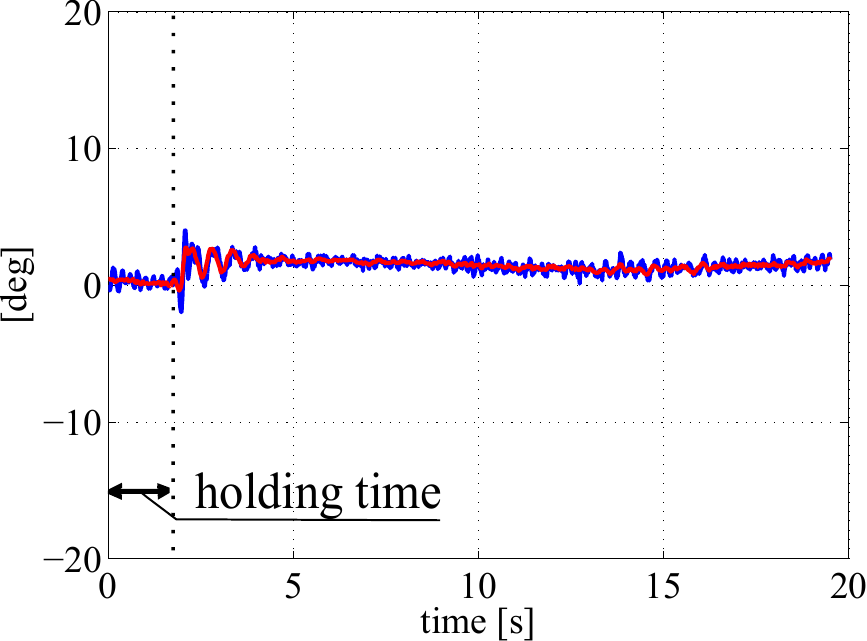}}\label{fig:theta_manreg}}\\
		\vspace{-0.1cm}		
		\subfloat[\emph{Thrust} $f(t)$] 
		{\hspace{-0.1cm}\includegraphics[width=3.9cm]{{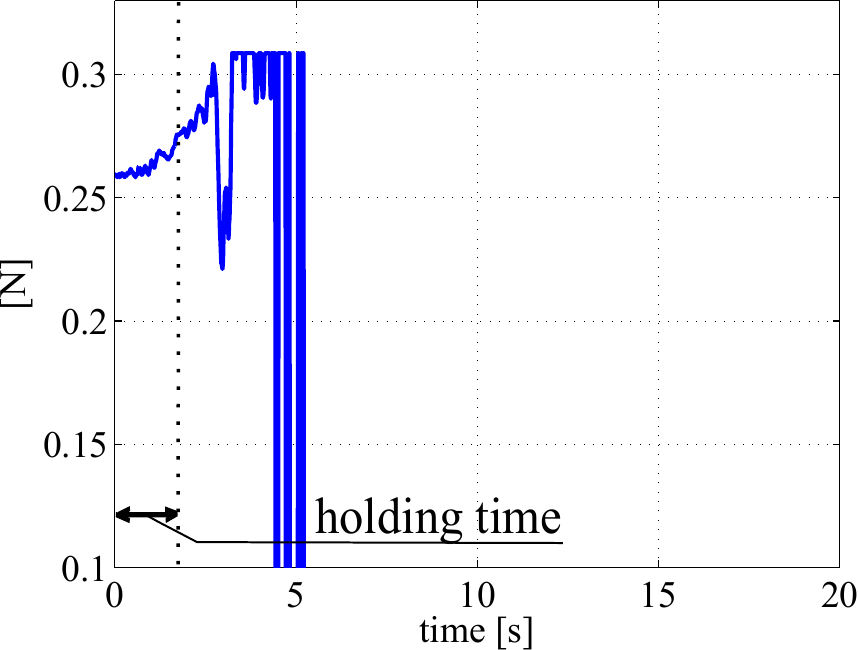}}\label{fig:f_track}
			\hspace{0.4cm}\includegraphics[width=3.9cm]{{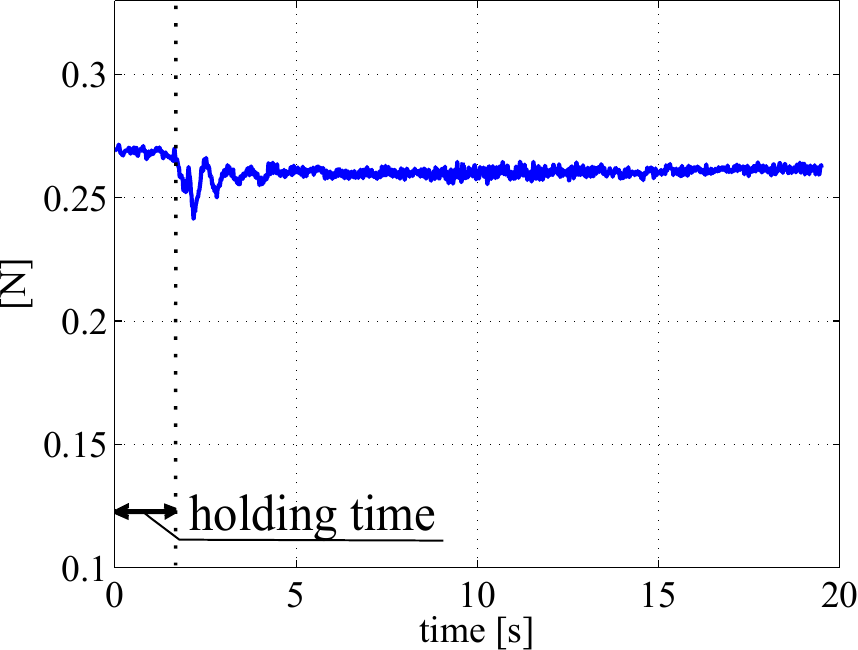}}\label{fig:f_manreg}}
		\caption{\emph{Trajectory Tracking} (left column) vs \emph{Maneuver Regulation} (right column).}
	\end{center}
\end{figure}
Let us first analyze the experiment in which the trajectory tracking controller
is adopted (left column). During the holding phase,
the desired state ``keeps moving" thus making the tracking error increase. When
the quadrotor is released, the desired position is ``far" from the actual
position, as depicted in Figure \ref{fig:xy_traj_track} on the left. The vehicle
attempts to quickly catch up the reference and this results into the foreseen
undesired phenomena.  A poor tracking of the desired path is realized: the
quadrotor does not track the circular arc, but chooses a shortest path to catch
up the position reference.  Moreover, the velocity (Figure \ref{fig:vel_track},
left) reaches a peak of more than $4.0$ m/s and the thrust (Figure
\ref{fig:f_track}, left) increases and saturates at the value of $0.31$ N. This
behavior causes an instability, as it can be seen from the angles depicted in
Figures~\ref{fig:phi_track} and~\ref{fig:theta_track} (left).  The vehicle is not
able to recover a controlled motion along the circle and finally falls down.
This dangerous behavior is avoided when using the maneuver regulation approach,
as shown in Figure 2 (right column).  While the quadrotor is constrained,
the reference state is suitably chosen according to the actual quadrotor
position. Since the reference position is selected as the one on the desired
path at minimum distance from the actual position, the position error does not
increase. As a consequence, when the quadrotor is released, the maneuver is
``smoothly regulated" thus converging to the desired trajectory. Moreover, as it
can be noticed in Figure \ref{fig:vel_manreg} (right), there is just a small
velocity overshoot (with a peak of less than $0.3$ m/s) due to a constant
velocity error during the constrained phase. The roll and pitch angles (Figures
\ref{fig:phi_manreg} and \ref{fig:theta_manreg}, right) closely follow the
reference, and the thrust (Figure \ref{fig:f_manreg}, right) does not
increase.

\begin{figure}[htbp]%[ht!]
	\begin{center}
		\hspace{-0.3cm}
		\subfloat[\emph{Path} $(p_1,p_2)$: desired position (red), actual position (blue).]
		{	\includegraphics[width=4.1cm]{{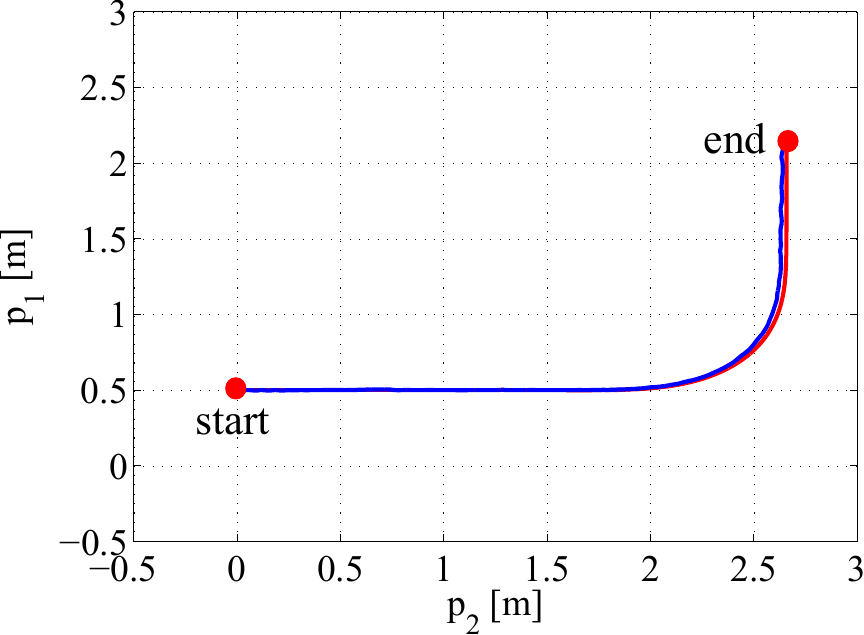}}\label{fig:xy_traj_fric}}
		\subfloat[\emph{Thrust} $f(t)$] 
		{\hspace{0.1cm}\includegraphics[width=4.0cm]{{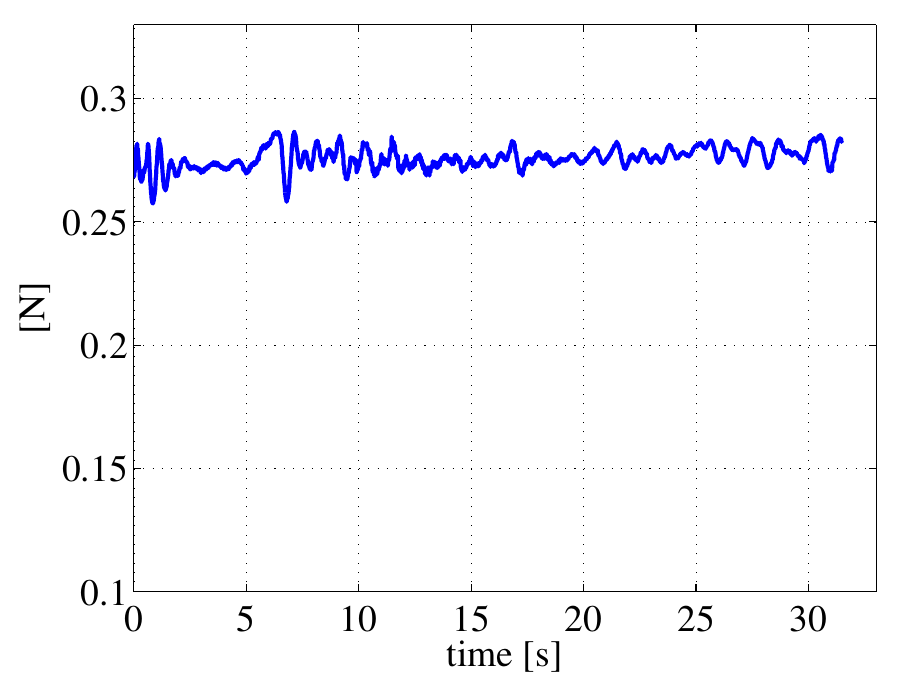}}\label{fig:f_fric}}\\
\subfloat[\emph{Positions}: $p_1(t)$ (magenta), $p_2(t)$ (black), $p_3(t)$ (green); dashed lines are desired positions, while solid lines are actual positions.]
{\includegraphics[width=4.0cm]{{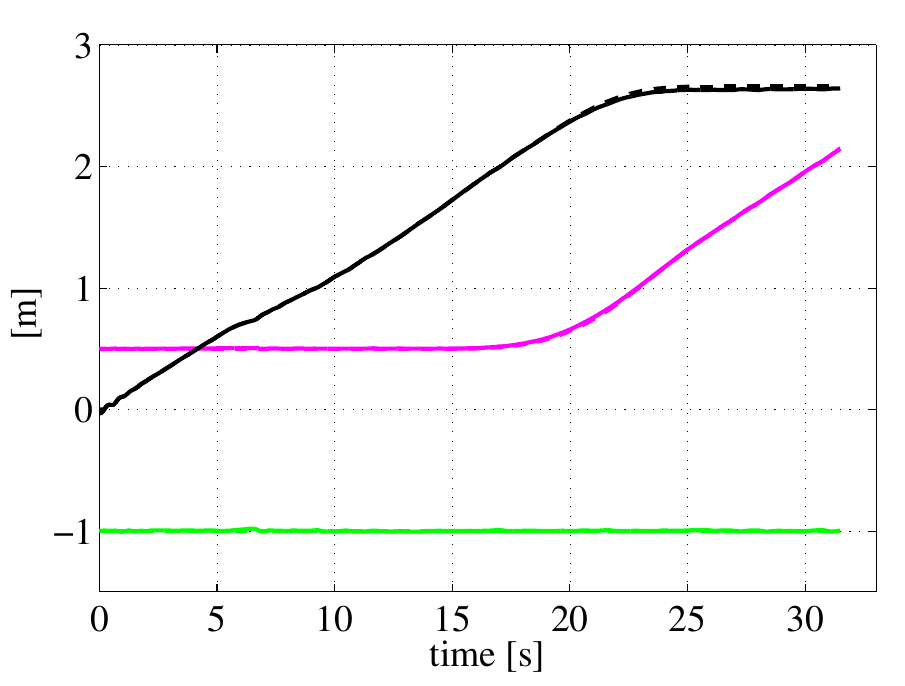}}\label{fig:p1p2p3_fric}}	\hspace{0.4cm}
\subfloat[\emph{Velocity norm} $\text{v}(t)$: desired velocity (red), actual velocity (blue).]{
	\hspace{-0.3cm}\includegraphics[width=4.0cm]{{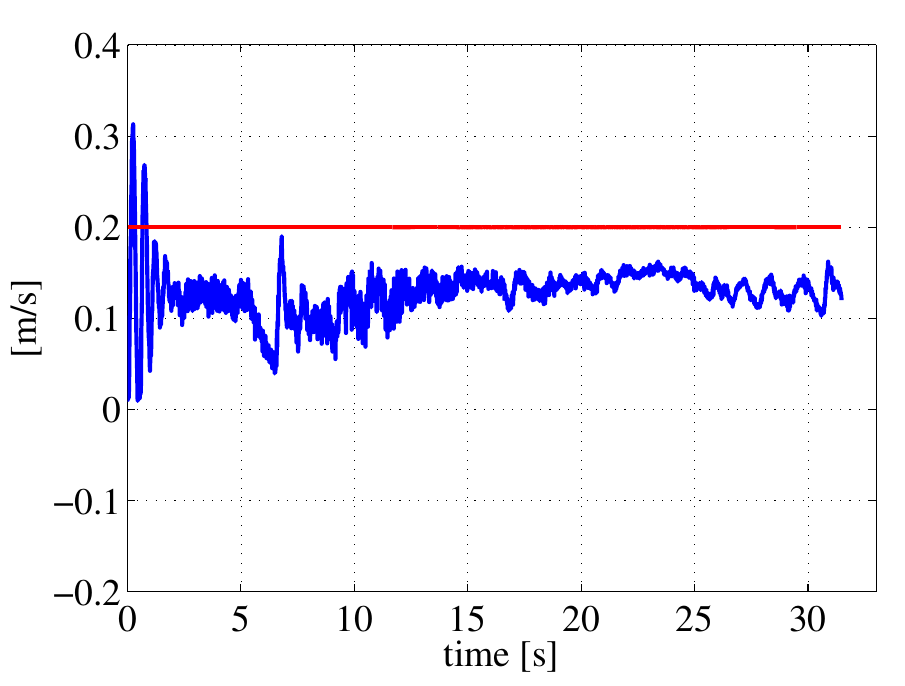}}\label{fig:vel_fric}}
\\
		\subfloat[\emph{Roll angle} $\varphi(t)$: roll angle command (red), actual angle (blue).]
		{\hspace{0.2cm}\includegraphics[width=4.0cm]{{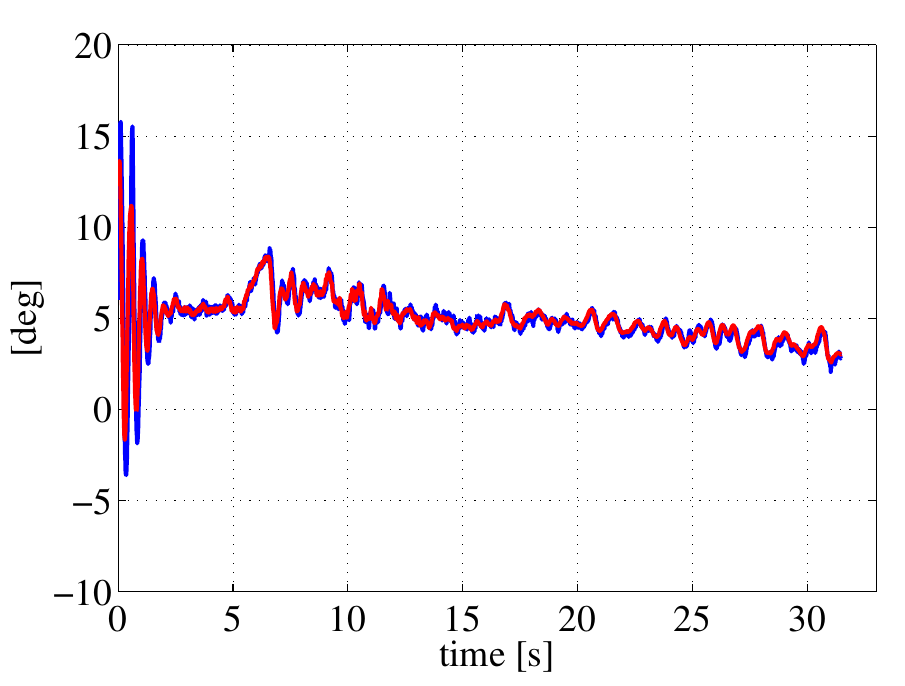}}\label{fig:phi_fric}} \hspace{0.2cm}
		\subfloat[\emph{Pitch angle} $\theta(t)$: pitch angle command (red), actual angle (blue).]
		{\includegraphics[width=4.0cm]{{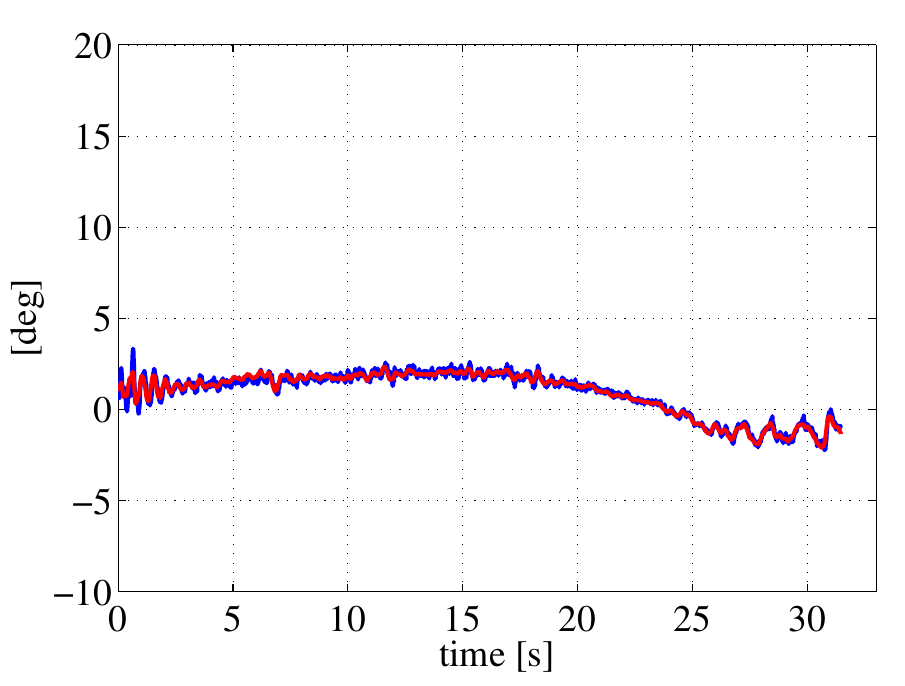}}\label{fig:theta_fric}}\\
		\subfloat[\emph{Experimental scenario} ] 
		{\includegraphics[width=3.9cm]{{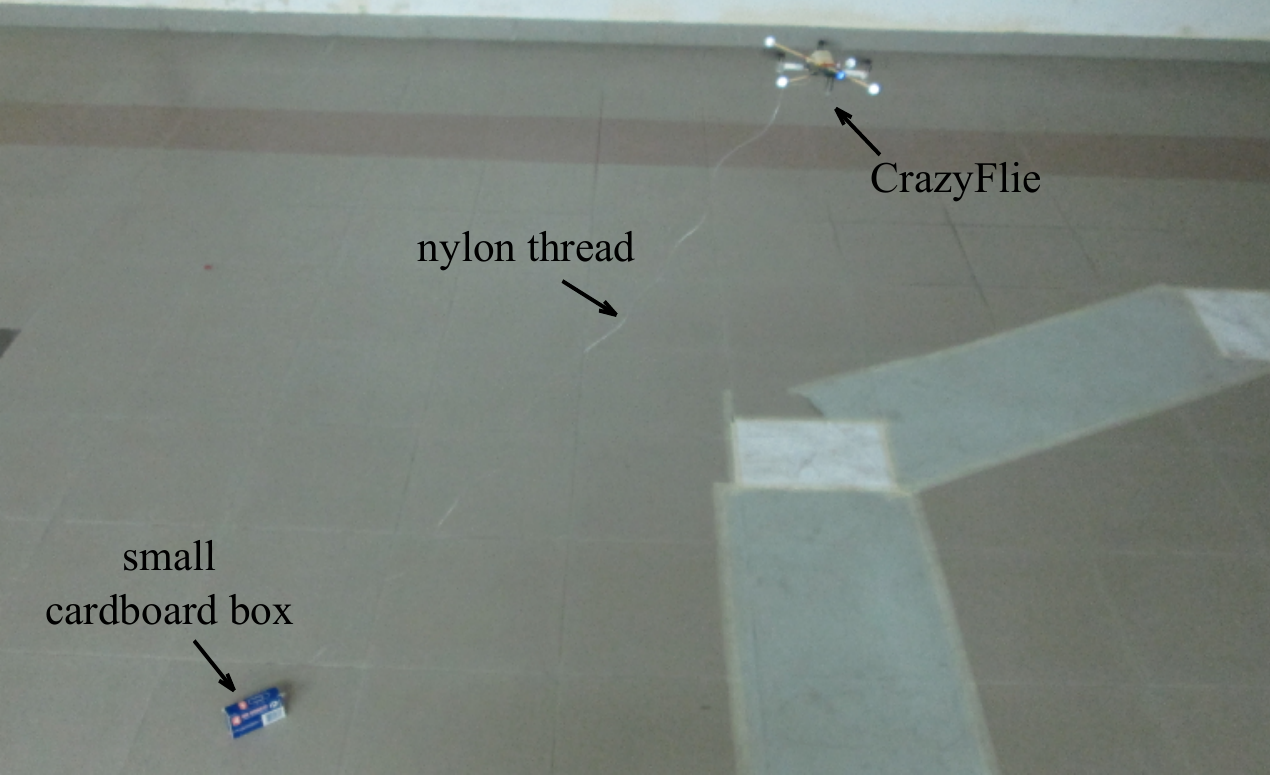}}\label{fig:picture1_fric}}
		\caption{Maneuver regulation controller: the vehicle performs a 90 degree turn while dragging a small payload.}
	\end{center}
\end{figure}

In order to further test the robustness of the maneuver regulation scheme under
disturbances, we perform a second experiment.  We choose, as desired trajectory,
a 90 degree turn with reference speed norm $\text{v}_d = 0.2 \; \text{m/s}$ and
yaw angle $\psi_d = 0$.  The quadrotor, controlled using our maneuver regulation
control law, is forced to execute the task when linked to a small
cardboard box through a nylon thread.
Corresponding results are depicted in Figure 3.
During a take off maneuver, the vehicle reaches the position
$\pmb{p}_d(0) = [0.5 \;\; 0.0 \; -1.0 ]^T \text{m}$: the quadrotor is linked to
the payload through the thread, but there is no traction through the cable
during this phase.
After the take off phase, we switch to the desired motion task. As soon as the
quadrotor starts moving closely to the desired trajectory, it slows down,
affected by the presence of the payload. The vehicle drags the payload through
the thread during all its motion. As a consequence, the motion of the vehicle is
decelerated with respect to the desired velocity reference, as shown in Figure
\ref{fig:vel_fric}. Nevertheless, the maneuver regulation controller is still
able to stabilize the vehicle, which closely follows the desired path, as
depicted in Figure \ref{fig:xy_traj_fric}.
% % % % % % % % % % % % % % % % % % % % % % % % % % % % % %
% % % % % % % % % % % % % % % % % % % % % % % % % % % % % % 

\section{Conclusions}
We have presented an easy-to-design maneuver regulation control strategy for
VTOL UAVs obtained by means of a trajectory tracking control
robustification. 
Specifically, we have designed a scheme in which the VTOL is required to stay on
a given path with a desired velocity profile, rather than catching up a desired
time-parametrized state. Since the maneuver regulation controller is derived
from a trajectory tracking control scheme, it inherits its properties (while
gaining the robustness of maneuver regulation) and does not need a new (possibly
time-consuming) parameter tuning.
To demonstrate the appealing features of the proposed controller, we have run
experimental tests on a 
nano-quadrotor. In these experiments, 
we have highlighted the robustness of maneuver regulation with respect to trajectory
tracking and demonstrated the correctness of 
our approach.

\addtolength{\textheight}{-12cm}   

\bibliographystyle{IEEEtran}
\bibliography{./bibAlias,./bibliography}

\end{document}